\documentclass[12pt]{article}
\usepackage[usenames]{color}
\usepackage{amsmath}
\usepackage{times}
\usepackage{graphicx}
\usepackage{multirow}
\usepackage[authoryear]{natbib}
\usepackage{rotating}
\usepackage{bbm}
\usepackage{latexsym}

\usepackage{listings}

\newcommand{\captionfonts}{\normalsize}

\makeatletter  
\long\def\@makecaption#1#2{%
  \vskip\abovecaptionskip
  \sbox\@tempboxa{{\captionfonts #1: #2}}%
  \ifdim \wd\@tempboxa >\hsize
    {\captionfonts #1: #2\par}
  \else
    \hbox to\hsize{\hfil\box\@tempboxa\hfil}%
  \fi
  \vskip\belowcaptionskip}
\makeatother   

\begin{document}
\hspace{13.9cm}1

\ \vspace{20mm}\\

{\LARGE Deep Big Simple Neural Nets Excel on Handwritten Digit Recognition}
\ \\
\\
{\bf \large Dan Claudiu Cire\c{s}an$^{\displaystyle 1, \displaystyle 2}$,} \\
{\bf \large Ueli Meier$^{\displaystyle 1, \displaystyle 2}$,} \\
{\bf \large Luca Maria Gambardella$^{\displaystyle 1, \displaystyle 2}$,} \\
{\bf \large J\"{u}rgen Schmidhuber$^{\displaystyle 1, \displaystyle 2}$}\\	
{$^{\displaystyle 1}$IDSIA, Galleria 2, 6928 Manno-Lugano, Switzerland.}\\
{$^{\displaystyle 2}$University of Lugano \& SUPSI, Switzerland.}\\
%

 \footnote{http://yann.lecun.com/exdb/mnist/}, BP (back-propagation).

\thispagestyle{empty}
\markboth{}{NC instructions}
\ \vspace{-0mm}\\
%
\begin{center} {\bf Abstract} \end{center}

Good old on-line back-propagation for plain multi-layer perceptrons yields a very low 0.35\% error rate on the famous MNIST handwritten digits benchmark. All we need to achieve this best result so far are many hidden layers, many neurons per layer, numerous deformed training images, and graphics cards to greatly speed up learning.

\section{Introduction}

Automatic handwriting recognition is of great academic and commercial interest. Current algorithms are already pretty good at learning to recognize handwritten digits. Post offices use them to sort letters; banks use them to read personal checks. MNIST \citep{Lecun:98} is the most widely used benchmark for isolated handwritten digit recognition. More than a decade ago, artificial neural networks called Multilayer Perceptrons or MLPs \citep{Werbos:74,LeCun:85,Rumelhart:86} were among the first classifiers tested on MNIST. Most had few layers or few artificial neurons (units) per layer \citep{Lecun:98}, but apparently back then they were the biggest feasible MLPs, trained when CPU cores were at least 20 times slower than today. A more recent MLP with a single hidden layer of 800 units achieved 0.70\% error \citep{Simard:03}. However, more complex methods listed on the MNIST web page always seemed to outperform MLPs, and the general trend went towards more and more complex variants of Support Vector Machines or SVMs \citep{Decoste:02} and combinations of NNs and SVMs \citep{Lauer:07} etc. Convolutional neural networks (CNNs) achieved a record-breaking 0.40\% error rate \citep{Simard:03}, using novel elastic training image deformations. Recent methods pre-train each hidden CNN layer one by one in an unsupervised fashion (this seems promising especially for small training sets), then use supervised learning to achieve 0.39\% error rate \citep{Ranzato:06, Ranzato:07}. The biggest MLP so far \citep{SalHinton:07} also was pre-trained without supervision then piped its output into another classifier to achieve an error of 1\% without domain-specific knowledge.

Are all these complexifications of plain MLPs really necessary? Can't one simply train really big plain MLPs on MNIST? Why is there no literature on this? One reason is that at first glance deep MLPs do not seem to work better than shallow networks \citep{Bengio:06}. Training them is hard as back-propagated gradients quickly vanish exponentially in the number of layers \citep{Hochreiter:91,Hochreiter:01,Hinton:07}, just like in the first recurrent neural networks \citep{Hochreiter:97}. Indeed, previous deep networks successfully trained with back-propagation (BP) either had few free parameters due to weight-sharing \citep[e.g.][]{Lecun:98,Simard:03} or used unsupervised, layer-wise pre-training \citep[e.g.][]{Bengio:06,Ranzato:06}.  But is it really true that deep BP-MLPs do not work at all, or do they just need more training time? How to test this? Unfortunately, on-line BP for hundreds/thousands of epochs on large MLPs may take weeks or months on standard serial computers. But can't one parallelize it? Well, on computer clusters this is hard due to communication latencies between individual computers. Multi-threading on a multi-core processor is not easy either. We may speed up BP using SSE (Streaming Single Instruction, Multiple Data Extensions), either manually, or by setting appropriate compiler flags. The maximum theoretical speedup under single precision floating-point, however, is four, which is not enough. And MNIST is large - its 60,000 images take almost 50MB, too much to fit in the L2/L3 cache of any current processor. This requires to continually access data in considerably slower RAM. To summarize, currently it is next to impossible to train big MLPs on CPUs.

We will show how to overcome all these problems by training large, deep MLPs on graphics cards.

\section{Data}

MNIST consists of two datasets, one for training (60,000 images) and one for testing (10,000 images). Many studies divide the training set into two sets consisting of 50,000 images for training and 10,000 for validation. Our network is trained on slightly deformed images, continually generated in on-line fashion; hence we may use the whole un-deformed training set for validation, without wasting training images. Pixel intensities of the original gray scale images range from 0 (background) to 255 (max foreground intensity). $28\times28=784$ pixels per image get mapped to real values $\frac{pixel~intensity}{127.5}-1.0$  in $[-1.0,1.0]$, and are fed into the NN input layer.

\section{Architectures}

We train 5 MLPs with 2 to 9 hidden layers and varying numbers of hidden units. Mostly but not always the number of hidden units per layer decreases towards the output layer (Table~\ref{Table:results}). There are 1.34 to 12.11 million free parameters (or weights, or synapses).

We use standard on-line BP \citep[e.g.][pages 744-748]{BOOK_AI}, without momentum, but with a variable learning rate that shrinks by a multiplicative constant after each epoch, from $10^{-3}$ down to $10^{-6}$. Weights are initialized with a uniform random distribution in $[-0.05,0.05]$. Each neuron's activation function is a scaled hyperbolic tangent: $y(a)=A\tanh{Ba}$, where $A=1.7159$ and $B=0.6666$ \citep{Lecun:98}. 

\section{Deforming images to get more training instances}

So far, the best results on MNIST were obtained by deforming training images, thus greatly increasing their number. This allows for training networks with many weights, making them insensitive to in-class variability. We combine affine (rotation, scaling and horizontal shearing) and elastic deformations, characterized by the following real-valued parameters:
\begin{itemize}
\item $\sigma$ and $\alpha$: for elastic distortions emulating uncontrolled oscillations of hand muscles \citep[for details see][]{Simard:03};
\item $\beta$: a random angle from $[-\beta,+\beta]$ describes either rotation or horizontal shearing. In case of shearing, $\tan\beta$ defines the ratio between horizontal displacement and image height;
\item $\gamma_x$, $\gamma_y$: for horizontal and vertical scaling, randomly selected from $[1-\gamma/100,1+\gamma/100]$.
\end{itemize}

At the beginning of every epoch the entire MNIST training set gets deformed. Initial experiments with small networks suggested the following deformation parameters: $\sigma=5.0-6.0$, $\alpha=36.0-38.0$, $\gamma=15-20$. Since digits 1 and 7 are similar they get rotated/sheared less ($\beta=7.5^\circ$) than other digits ($\beta=15.0^\circ$).

\section{Results}

All simulations were performed on a computer with a Core2 Quad 9450 2.66GHz processor, 3GB of RAM, and a GTX280 graphics card. The GPU accelerates the deformation routine by a factor of 10 (only elastic deformations are GPU-optimized); the forward propagation (FP) and BP routines are sped up by a factor of 40. Implementation details can be found in the Appendix. We pick the trained MLP with the lowest validation error, and evaluate it on the MNIST test set. Results are summarized in Table~\ref{Table:results}.

\begin{figure}[ht]
\hfill
\begin{center}
\includegraphics[width=0.75\textwidth]{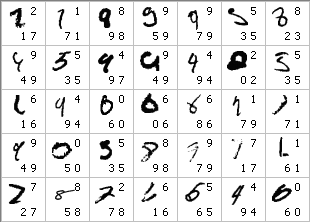}
\end{center}
\caption{The 35 miss-classified digits of the best network from Table~\ref{Table:results}, together with the two most likely predictions (bottom, from left to right) and the correct label according to MNIST (top, right).}
\label{Fig:errors}
\end{figure}

Most remarkably, the best network has an error rate of only 0.35\% (35 out of 10,000 digits). This is significantly better than the best previously published results, namely, 0.39\% by \cite{Ranzato:06} and 0.40\% by \cite{Simard:03}, both obtained by more complex methods. The 35 misclassified digits are shown in Figure~\ref{Fig:errors}. Many of them are ambiguous and/or uncharacteristic, with obviously missing parts or strange strokes etc. Interestingly, the second guess of the network is correct for 30 out of the 35 misclassified digits.

The best test error of this MLP is even lower (0.32\%) and may be viewed as the maximum capacity of the network. Performance clearly profits from adding hidden layers and more units per layer.  For example, network 5 has more but smaller hidden layers than network 4 (Table~\ref{Table:results}).

Networks with up to 12 million weights can successfully be trained by plain gradient descent to achieve test errors below 1\% after 20-30 epochs in less than 2 hours of training. How can networks with so many parameters generalize well on the unseen test set? Answer: the continual deformations of the training set generate a virtually infinite supply of training examples, and the network rarely sees any training image twice.

\begin{table}[h]
	\footnotesize
	\caption{Error rates on MNIST test set.}
	\label{Table:results}
	\vspace{8pt}
	\centering
  \begin{tabular}{c|c|ccrr}
    ID	&	architecture 													& test error for 			& best test		& simulation	&	weights		\\
    		&	(number of neurons in each layer)			&	best validation [\%]	&	error [\%]	&	 time [h]		&	[milions]	\\
    \hline \hline
    1		&	1000, 500, 10 												&	\textbf{0.49}				&	0.44				&	23.4				&	1.34			\\
    2		&	1500, 1000, 500, 10 									&	\textbf{0.46}				&	0.40				& 44.2				&	3.26			\\
    3		&	2000, 1500, 1000, 500, 10 						&	\textbf{0.41}				&	0.39				&	66.7				&	6.69			\\
    4		&	2500, 2000, 1500, 1000, 500, 10 			&	\textbf{0.35}				&	0.32				&	114.5				&	12.11			\\
    \hline
    5		&	9 $\times$ 1000, 10  									&	\textbf{0.44}				&	0.43				&	107.7				&	8.86			\\
  \end{tabular}
\end{table}

\section*{Conclusion}

In recent decades the amount of raw computing power per Euro has grown by a factor of 100-1000 per decade. Our results show that this ongoing hardware progress may be more important than advances in algorithms and software (although the future will belong to methods combining the best of both worlds). Current graphics cards (GPUs) are already more than 40 times faster than standard microprocessors when it comes to training big and deep neural networks by the ancient algorithm, on-line back-propagation (weight update rate up to $5\times10^9/s$, and more than $10^{15}$ per trained network).  On the very competitive MNIST handwriting benchmark, single precision floating-point GPU-based neural nets surpass all previously reported results, including those obtained by much more complex methods involving specialized architectures, unsupervised pre-training, combinations of machine learning classifiers etc. Training sets of sufficient size are obtained by appropriately deforming images. Of course, the approach is not limited to handwriting, and obviously holds great promise for many visual and other pattern recognition problems.

\subsection*{Acknowledgments}

Part of this work got started when Dan Cire\c{s}an was a PhD student at University ''Politehnica'' of Timi\c{s}oara. He would like to thank his former PhD advisor, \c{S}tefan Holban, for his guidance, and R\u{a}zvan Mo\c{s}incat for providing a CPU framework for MNIST. This work was supported by Swiss CTI, Commission for Technology and Innovation, Project n. 9688.1 IFF: Intelligent Fill in Form, and by Lifeware S.A. L.M.G. and J.S. wrote a grant proposal for this work, acquired competitive funding for it, and supervised it. C.D.C. wrote GPU-optimised code. C.D.C. and U.M. debugged it. C.D.C. designed and performed the experiments. C.D.C., U.M., L.M.G. and J.S. wrote the paper.

\section*{Appendix - GPU implementation}

\subsection*{Graphics Processing Unit}
Until 2007 the only way to program a GPU was to translate the problem-solving algorithm into a set of graphical operations. Despite being hard to code and difficult to debug, several GPU-based NN implementations were developed when GPUs became faster than CPUs. Two layer MLPs \citep{GPU2005} and CNNs \citep{GPU2006} have been previously implemented on GPUs. Although speedups were relatively modest, these studies showed how GPUs can be used for machine learning. More recent GPU-based CNNs trained in batch mode are two orders of magnitude faster than CPU-based CNNs \citep{Scherer:09}.

In 2007, NVIDIA developed the first version of CUDA (Compute Unified Device Architecture), a C-like general programming language. GPU speed and memory bandwidth are vastly superior to those of CPUs, and crucial for fast MLP implementations. To fully understand our algorithm in terms of GPU / CUDA, please visit the NVIDIA website \citep{NVIDIA2009}. According to CUDA terminology, the CPU is called \textit{\textbf{host}} and the graphics card \textit{\textbf{device}} or \textit{\textbf{GPU}}.

\subsection*{Deformations}
It takes 93 CPU seconds to deform the 60,000 MNIST training images, most of them (87) for elastic distortions. Only the most time-consuming part of the latter -- convolution with a gaussian kernel \citep{Simard:03} -- is ported to the GPU. The MNIST training set is split into 600 sequentially processed batches. MNIST digits are scaled from the original $28\times28$ pixels to $29\times29$ pixels, to get a proper center, which simplifies convolution. An image grid has 290 $\times$ 290 cells, zero-padded to 300 $\times$ 300, thus avoiding margin effects when applying a gaussian convolution kernel of size $21\times21$.
\begin{lstlisting}[caption=Convolution Kernel for elastic distortion.,
  label=Listing:Conv,
  float=t]
__global__ void ConvolveField_optimized(float *randomfield, int width, int height, float *kernel, float *outputfield, float elasticScale){
	float sum=0;
	const int stride_k=GET_STRIDE(GAUSSIAN_FIELD_SIZE,pitch_x>>2);	//stride for gaussian kernel
	__shared__ float K[GAUSSIAN_FIELD_SIZE][stride_k];				//kernel (21 x 32 values)
	__shared__ float R[GAUSSIAN_FIELD_SIZE+9][GAUSSIAN_FIELD_SIZE];	//random field (30 x 21 values)
	__shared__ float s[10][GAUSSIAN_FIELD_SIZE];					//partial sums (10 x 21 values)
	int stride_in=GET_STRIDE(width,pitch_x>>2);						//random field stride as a multiple of 32
	int stride_out=GET_STRIDE(width-GAUSSIAN_FIELD_SIZE+1,pitch_x>>2);	//output stride as a multiple of 32

	//loading gaussian kernel into K (21 x 21 values)
	K[ 0+threadIdx.y][threadIdx.x] = kernel[( 0+threadIdx.y)*stride_k + threadIdx.x];//rows 0..9
	K[10+threadIdx.y][threadIdx.x] = kernel[(10+threadIdx.y)*stride_k + threadIdx.x];//rows 10..19
	if(threadIdx.y==0)
		K[20+threadIdx.y][threadIdx.x] = kernel[(20+threadIdx.y)*stride_k + threadIdx.x];//row 20

	//loading randomfield into R
	//0..9 x 21 values
	R[ 0+threadIdx.y][threadIdx.x] = randomfield[(10*blockIdx.y+ 0+threadIdx.y)*stride_in + blockIdx.x + threadIdx.x];
	//10..19 x 21 values
	R[10+threadIdx.y][threadIdx.x] = randomfield[(10*blockIdx.y+10+threadIdx.y)*stride_in + blockIdx.x + threadIdx.x];
	//20..29 x 21 values
	R[20+threadIdx.y][threadIdx.x] = randomfield[(10*blockIdx.y+20+threadIdx.y)*stride_in + blockIdx.x + threadIdx.x];
	__syncthreads();	//wait until everything is read into shared memory

	//computing partial sums
	#pragma unroll 21	//GAUSSIAN_FIELD_SIZE
	for(int i=0;i<GAUSSIAN_FIELD_SIZE;i++)
		sum += R[threadIdx.y + i][threadIdx.x] * K[i][threadIdx.x];
	s[threadIdx.y][threadIdx.x]=sum;
	__syncthreads();

	if(threadIdx.x==0){	//the first column of threads compute the final values of the convolutions
		#pragma unroll 20//GAUSSIAN_FIELD_SIZE-1	
		for(int i=1;i<GAUSSIAN_FIELD_SIZE;i++) sum+=s[threadIdx.y][i];
		outputfield[(blockIdx.y*10+threadIdx.y)*stride_out + blockIdx.x] = sum * elasticScale;
	}
}\end{lstlisting}

Our GPU program groups many threads into a block, where they share the same gaussian kernel and parts of the random field. The blocks contain 21 (the kernel size) $\times$10 threads, each computing a vertical strip of the convolution operation (Listing~\ref{Listing:Conv}).

Generating the elastic displacement field takes only 3 seconds. Deforming the whole training set is more than 10 times faster, taking 9 instead of the original 93 seconds. Further optimization would be possible by porting all deformations onto the GPU, and by using the hardware's interpolation capabilities to perform the final bilinear interpolation. We omitted this since deformations are already pretty fast (deforming all images of one epoch takes only 5-15 \% of total computation time, depending on MLP size).

\section*{Training algorithm}
We closely follow the standard BP algorithm \citep[e.g.][pages 744-748]{BOOK_AI}, except that BP of deltas and weight updates are disentangled and performed sequentially. This allows for more parallelism within each routine.

\subsection*{Forward propagation}
\begin{figure}[ht]
\hfill
\begin{center}
\includegraphics[width=\textwidth]{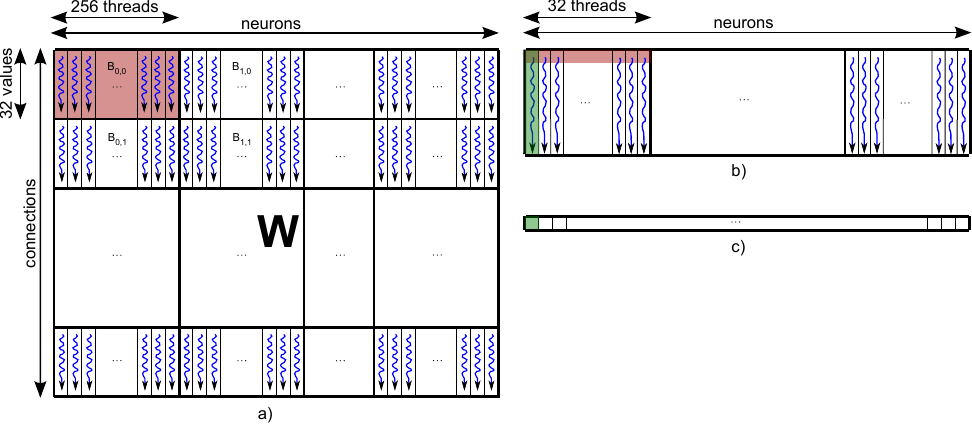}
\end{center}
\caption{Forward propagation: a) mapping of kernel 1 grid onto the padded weight matrix; b) mapping the kernel 2 grid onto the partial dot products matrix; c) output of forward propagation.}
\label{Fig:FP}
\end{figure}

The algorithm is divided into two kernels. The weight matrix $W$ is partitioned as illustrated in Figure~\ref{Fig:FP}. 

\textbf{Kernel 1}

Each block has 256 threads (Figure~\ref{Fig:FP}a), each computing a partial dot product of 32 component vectors. The dot products are stored in a temporary matrix (Figure~\ref{Fig:FP}b). This kernel has a very high throughput: average memory bandwidth is 115GB/s. This is possible because many relatively small blocks keep the GPU busy. Each block uses shared memory for storing the previous layer activations, which are simultaneously read by the first 32 threads of each block and then used by all 256 threads. After thread synchronization, the partial dot products are computed in parallel (Listing~\ref{Listing:FP}). The number of instructions is kept to a minimum by pre-computing all common index parts.

\begin{lstlisting}[caption=Forward propagation kernels.,
  label=Listing:FP,
  float=ht]
__global__ void MLP_FP_reduction_Kernel1(float *prevLN, float *W, float *partialsum, unsigned int neurons, unsigned int prevneurons){
	const int threads=256;
	const int stride=GET_STRIDE(neurons,pitch_x>>2);	//horizontal stride of W matrix
	int X=blockIdx.x*threads + threadIdx.x;		//precomputing expressions
	int Y=X+stride*blockIdx.y;
	int Z=blockIdx.y*pitch_y*stride + X;
	float sum=0.0f;
	__shared__ float output[pitch_y];		
	if(blockIdx.y==0)
		if(threadIdx.x==0) output[0]=1.0f;
		else if(threadIdx.x<pitch_y)	//there are only 32 values to read and 128 threads
			output[threadIdx.x] = threadIdx.x-1<prevneurons ? prevLN[threadIdx.x-1] : 0.0f;
			else;
	else if(threadIdx.x<pitch_y)	//there are only 32 values to read and 128 threads
			output[threadIdx.x] = blockIdx.y*pitch_y+threadIdx.x-1<prevneurons ? 
							prevLN[blockIdx.y*pitch_y+threadIdx.x-1] : 0.0f;
	     else;
	__syncthreads();
	if(X<neurons){//compute partial sums
		//#pragma unroll 32
		int size=0;
		if((blockIdx.y+1)*pitch_y>=prevneurons+1)
			size = prevneurons + 1 - blockIdx.y*pitch_y;
		else size=pitch_y;
		for (int ic=0; ic<size; ic++){
			sum += output[ic] * W[Z];
			Z+=stride;
		}
		partialsum[Y]=sum;	
	}
}

__global__ void MLP_FP_reduction_Kernel2(float *currLN, float *partialsum, unsigned int neurons, unsigned int size){
	float sum=0.0f;
	int idx = blockIdx.x*(pitch_x>>2) + threadIdx.x;		//precomputed index
	unsigned int stride = GET_STRIDE(neurons,pitch_x>>2);		//stride for partialsum matrix

	if(idx>=neurons)return;						//is this thread computing a true neuron?
	for (int i=0; i<size; i++) sum += partialsum[i*stride+idx];	//computing the final dot product
	currLN[idx] = SIGMOIDF(sum);					//applying activation
}

\end{lstlisting}

\textbf{Kernel 2}

The thread grid (Figure~\ref{Fig:FP}b) has only one row of blocks consisting of $warp$ threads, since each thread has to compute a complete dot product (Figure~\ref{Fig:FP}c) and then pipe it into the activation function. 
This kernel (Listing~\ref{Listing:FP}) is inefficient for layers with fewer than 
1024 incoming connections per neuron, especially  for the last layer which has only ten neurons, one for each digit. That is, its grid will have only one block, occupying only 3\% of the GTX280 GPU. 

\subsection*{Backward propagation}
This is similar to FP, but we need $W^T$ for coalesced access. Instead of transposing the matrix, the computations are performed on patches of data read from device memory into shared memory, similar to the optimized matrix transposition algorithm of \cite{Ref2009Transpose}. Shared memory access is much faster, without coalescing restrictions. Because we have to cope with layers of thousands of neurons, back-propagating deltas uses a reduction method implemented in two kernels communicating partial results via global memory (Listing~\ref{Listing:BP_deltas}).

\textbf{Kernel 1}
The bi-dimensional grid is divided into blocks of $warp$ (32) threads. The kernel starts by reading a patch of $32\times32$ values from W. The stride of the shared memory block is 33 ($warp+1$), thus avoiding all bank conflicts and significantly improving speed. Next, 32 input delta values are read and all memory locations that do not correspond to real neurons (because of vertical striding) are zero-padded to avoid branching in subsequent computations. The number of elements is fixed to $warp$ size, and the computing loop is unrolled for further speedups. Before finishing, each thread writes its own partial dot product to global memory. 

\begin{lstlisting}[caption=Backpropagating deltas kernels.,
  label=Listing:BP_deltas,
  float=ht]
  
__global__ void backPropagateDeltasFC_s2_A(float *indelta, float *weights, unsigned int ncon, unsigned int nrneur, float *partial){
	const int		px = pitch_x>>2;
	unsigned int	stride_x = GET_STRIDE(nrneur,px);
	unsigned int	stride_y = GET_STRIDE(ncon,pitch_y);
	float			outd = 0.0;
	int				idx = blockIdx.x*px+threadIdx.x;
	int				X = blockIdx.y*pitch_y*stride_x + idx;
	int				Y = threadIdx.x;
	__shared__ float w[32*33];				//pitch_y and px should be equal ! +1 to avoid bank conflict!
	__shared__ float id[px];				//input delta
		#pragma unroll 32	//read the weight patch in shared memory
	for(int i=0;i<pitch_y;i++){w[Y]=weights[X]; X+=stride_x; Y+=33;}
	//read the input delta patch in shared memory
	if(idx>=nrneur)	id[threadIdx.x]=0;	//a fake input delta for inexistent indelta
	else id[threadIdx.x]=indelta[idx];
	__syncthreads();	//not needed for block with warp number of threads: implicit synchronization
	#pragma unroll 32 //compute partial results
	for(int i=0;i<px;i++) outd+=w[threadIdx.x*33+i]*id[i];
	//write out the partial results
	partial[blockIdx.x*stride_y + blockIdx.y*pitch_y +  threadIdx.x] = outd;
}
__global__ void backPropagateDeltasFC_s2_B(float *outdelta,float *instates, unsigned int ncon, unsigned int nrneur, float *partial){
	int px=pitch_x>>2;
	unsigned int stride_x = GET_STRIDE(nrneur,px);
	unsigned int stride_y = GET_STRIDE(ncon,pitch_y);
	float outd = 0.0;
	int size=stride_x/px;
	int idx=blockIdx.x*pitch_y+threadIdx.x;
	if(idx==0);	//true only for block and thread 0
	else{
		for(int i=0;i<size;i++)
			outd+=partial[i*stride_y + idx];
		outdelta[idx-1] = outd * DSIGMOIDF(instates[idx-1]);	//-1 BIAS ...
	}
}
\end{lstlisting}

\textbf{Kernel 2}

This kernel completes BP of deltas by summing up partial deltas computed by the previous kernel.  It multiplies the final result by the derivative of the activation function applied to the current neuron's state, and writes the new delta to global memory.

\subsection*{Weight updating}
\begin{lstlisting}[caption=Weights adjustment kernel.,
  label=Listing:WA,
  float=t]
__global__ void adjustWeightsFC_s1(float *states,float *deltas, float *weights, float eta, unsigned int ncon, unsigned int nrneur){
	const int pitch_y=16;
	const int threads=256;
	unsigned int px = pitch_x >> 2;
	unsigned int stride_x = GET_STRIDE(nrneur,px);
	float etadeltak = eta*deltas[blockIdx.x*threads+threadIdx.x],t;
	int b=blockIdx.y*stride_x*pitch_y + threads*blockIdx.x + threadIdx.x;
	__shared__ float st[pitch_y];	//for states
	int cond1 = blockIdx.y || threadIdx.x;
	int cond2 = (blockIdx.y+1)*pitch_y<=ncon;
	int size = cond2 * pitch_y + !cond2 * (ncon%pitch_y);	
	if(threadIdx.x<pitch_y)	st[threadIdx.x] = cond1 * states[blockIdx.y*pitch_y + threadIdx.x - 1] + !cond1;
	__syncthreads();

	if (blockIdx.x*threads + threadIdx.x < nrneur){
		#pragma unroll 16
		for (int j=0; j<16; j++){
			t=weights[b];
			t-= etadeltak * st[j];
			weights[b]=t;
			b+=stride_x;}}
}
\end{lstlisting}

The algorithm (Listing~\ref{Listing:WA}) starts by reading the appropriate delta, and pre-computes all repetitive expressions. Then the first 16 threads read the states from global memory into shared memory. The ``bias neuron'' with constant activation 1.0 is dealt with by conditional statements, which could be avoided through expressions containing the conditions. Once threads are synchronized, each single thread updates 16 weights in a fixed unrolled loop.


\clearpage

\begin{thebibliography}{100}
\providecommand{\natexlab}[1]{#1}
\expandafter\ifx\csname urlstyle\endcsname\relax
  \providecommand{\doi}[1]{doi:\discretionary{}{}{}#1}\else
  \providecommand{\doi}{doi:\discretionary{}{}{}\begingroup
  \urlstyle{rm}\Url}\fi


\bibitem[{Bengio et~al.(2006)Bengio, Lamblin, Popovici \& Larochelle}]{Bengio:06}
Bengio, Y., Lamblin, P., Popovici, D. \& Larochelle, H. (2006).
\newblock Greedy layer-wise training of deep networks.
\newblock \emph{Advances in Neural Information Processing Systems 19 (NIPS'06)}, 153 -- 160.

\bibitem[{Chellapilla et~al.(2005)Chellapilla, K., Puri, S. \& Simard, P.}]{GPU2006}
Chellapilla, K., Puri, S. \& Simard, P. (2006).
\newblock High Performance Convolutional Neural Networks for Document Processing.
\newblock \emph{10th International Workshop on Frontiers in Handwriting Recognition}, 2006.

\bibitem[{Decoste \& Scholkopf(2002)Decoste \& Scholkopf}]{Decoste:02}
Decoste, D. \& Scholkopf, B. (2002).
\newblock Training Invariant Support Vector Machines.
\newblock \emph{Machine learning}, \emph{46}, 161 -- 190.

\bibitem[{Hinton(2007)}]{Hinton:07}
Hinton, G. (2007).
\newblock To recognize shapes, first learn to generate images.
\newblock \emph{Computational Neuroscience: Theoretical Insights into Brain Function.}, Elsevier.

\bibitem[{Hochreiter(1991)}]{Hochreiter:91}
Hochreiter, S. (1991).
\newblock Untersuchungen zu dynamischen neuronalen Netzen.
\newblock \emph{Diploma thesis, Institut f\"{u}r Informatik, Lehrstuhl Prof. Brauer, Technische Universit\"{a}t M\"{u}nchen}.

\bibitem[{Hochreiter et al.(2001)}]{Hochreiter:01}
Hochreiter, S., Bengio, Y., Frasconi, P. \& Schmidhuber, J. (2001).
\newblock Gradient flow in recurrent nets: the difficulty of learning long-term dependencies.
\newblock \emph{A Field Guide to Dynamical Recurrent Neural Networks}, IEEE Press.

\bibitem[{Hochreiter \& Schmidhuber(1997)}]{Hochreiter:97}
Hochreiter, S. \& Schmidhuber, J. (1997).
\newblock Long Short-Term Memory.
\newblock \emph{Neural Computation}, \emph{9}, 1735 -- 1780.

\bibitem[{Lauer et~al.(2007)Lauer, Suen \& Bloch (2007).
Authors(2008)Author1, Author2, \&  Author3}]{Lauer:07}
Lauer, F., Suen, C. \& Bloch, G. (2007).
\newblock A trainable feature extractor for handwritten digit recognition.
\newblock \emph{Pattern Recognition}, \emph{40}, 1816 -- 1824.

\bibitem[{LeCun(1985)}]{LeCun:85}
LeCun, Y. (1985).
\newblock Une proc\'{e}dure d'apprentissage pour r\'{e}seau \`{a} seuil asym\'{e}trique.
\newblock \emph{Proceedings of Cognitiva, Paris}, \emph{85}, 599 -- 604.

\bibitem[{LeCun et~al.(1998)LeCun, Bottou, Bengio, Haffner}]{Lecun:98}
LeCun, Y., Bottou, L., Bengio, Y. \& Haffner, P. (1998).
\newblock Gradient-Based Learning Applied to Document Recognition.
\newblock \emph{Proceedings of the IEEE}, \emph{86}, 309 -- 318.

\bibitem[{NVIDIA(2009)}]{NVIDIA2009}
NVIDIA. (2009).
\newblock NVIDIA CUDA. Reference Manual.
\newblock \emph{version 2.3}, 2009.

\bibitem[{Ranzato et~al.(2007), Ranzato, Huang, Boureau \& LeCun}]{Ranzato:07}
Ranzato, M., Huang, F., Boureau, Y. \& LeCun, Y.(2007).
\newblock Unsupervised Learning of Invariant Feature Hierarchies with Applications to Object Recognition.
\newblock \emph{Proc. Computer Vision and Pattern Recognition Conference (CVPR'07)}.

\bibitem[{Ranzato et~al.(2006), Ranzato, Poultney, Chopra \& LeCun}]{Ranzato:06}
Ranzato, M., Poultney, C., Chopra, S. \& LeCun, Y. (2006).
\newblock Efficient Learning of Sparse Representations with an Energy-Based Model.
\newblock \emph{Advances in Neural Information Processing Systems (NIPS 2006)}.

\bibitem[{Ruetsch \& Micikevicius(2009)Ruetsch \& Micikevicius}]{Ref2009Transpose}
Ruetsch, G. \& Micikevicius, P. (2009).
\newblock Optimizing Matrix Transpose in CUDA.
\newblock \emph{NVIDIA GPU Computing SDK}, 1 -- 24.

\bibitem[{Rumelhart et~al.(1986)Rumelhart, Hinton \& Williams}]{Rumelhart:86}
Rumelhart, D. E., Hinton, G. E. \& Williams, R. J. (1986).
\newblock Learning internal representations by error propagation.
\newblock \emph{Parallel Distributed Processing}, \emph{MIT Press}, \emph{1}, 318 -- 362.

\bibitem[{Russell \& Norvig(2002)Russell \& Norvig}]{BOOK_AI}
Russell, S. \& Norvig, P. (2002).
\newblock Artificial Intelligence: A Modern Approach (2nd Edition).
\newblock \emph{Prentice Hall}, \emph{ISBN-13: 978-01379039555}.

\bibitem[{Salakhutdinov et~al.(2007)Salakhutdinov, R. \& Hinton, G.}]{SalHinton:07}
Salakhutdinov, R. \& Hinton, G. (2007).
\newblock Learning a Nonlinear Embedding by Preserving Class Neighborhood Structure.
\newblock \emph{Proceedings of the International Conference on Artificial Intelligence and Statistics}, \emph{11}, 2007.

\bibitem[{Scherer \& Behnke(2009)}]{Scherer:09}
Scherer, D. \& Behnke, S. (2009).
\newblock Accelerating Large-scale Convolutional Neural Networks with Parallel Graphics Multiprocessors.
\newblock \emph{Proc. of NIPS 2009 Workshop on Large-Scale Machine Learning: Parallelism and Massive Datasets}.

\bibitem[{Simard et~al.(2003)Simard,Steinkraus,Platt}]{Simard:03}
Simard, P.Y., Steinkraus, D. \& Platt, J.C. (2003).
\newblock Best Practices for Convolutional Neural Networks Applied to Visual Document Analysis.
\newblock \emph{Intl. Conf. Document Analysis and Recognition}, 958 -- 962.

\bibitem[{Steinkraus et~al.(2005)Steinkraus, D., Buck, I. \& Simard, P.}]{GPU2005}
Steinkraus, D., Buck, I. \& Simard, P.Y. (2005).
\newblock GPUs for Machine Learning Algorithms.
\newblock \emph{ICDAR 2005}, \emph{2}, 1115 -- 1120.

\bibitem[{Werbos(1974)}]{Werbos:74}
Werbos, P.J. (1974).
\newblock Beyond Regression: New Tools for Prediction and Analysis in the Behavioral Sciences.
\newblock \emph{PhD thesis, Harvard University.}

%

\end{thebibliography}
\end{document}